\documentclass[11pt,a4paper]{article}
\usepackage[hyperref]{emnlp2020}
\usepackage{times}
\usepackage{latexsym}
\usepackage{pgfplots}
\usepackage{graphicx}
\usepackage{graphics}
\usepackage{textcomp}
\usepackage{tikz}
\usepackage{multirow}
\usepackage{enumitem}
\newcommand{\specialcell}[2][l]{%
  \begin{tabular}[#1]{@{}l@{}}#2\end{tabular}}
\usepackage[T1]{fontenc}
\usepackage[utf8]{inputenc}

\pgfplotsset{compat=1.16}

\usepackage{microtype}

\aclfinalcopy 

\title{FQuAD2.0: French Question Answering and knowing that you know nothing}

\author{Quentin Heinrich, Gautier Viaud, Wacim Belblidia \\
  Illuin Technology \\
  Paris, France \\
  \texttt{\{quentin, gautier, wacim\}@illuin.tech} \\
}

\date{\today}

\begin{document}
\maketitle
\begin{abstract}
Question Answering, including Reading Comprehension, is one of the NLP research areas that has seen significant scientific breakthroughs over the past few years, thanks to the concomitant advances in Language Modeling.
Most of these breakthroughs, however, are centered on the English language.
In 2020, as a first strong initiative to bridge the gap to the French language, Illuin Technology introduced FQuAD1.1, a French Native Reading Comprehension dataset composed of 60,000+ questions and answers samples extracted from Wikipedia articles.
Nonetheless, Question Answering models trained on this dataset have a major drawback: they are not able to predict when a given question has no answer in the paragraph of interest, therefore making unreliable predictions in various industrial use-cases.
In the present work, we introduce FQuAD2.0, which extends FQuAD with 17,000+ unanswerable questions, annotated adversarially, in order to be similar to answerable ones.
This new dataset, comprising a total of almost 80,000 questions, makes it possible to train French Question Answering models with the ability of distinguishing unanswerable questions from answerable ones.
We benchmark several models with this dataset: our best model, a fine-tuned CamemBERT\textsubscript{LARGE}, achieves a F1 score of 82.3\% on this classification task, and a F1 score of 83\% on the Reading Comprehension task.

\end{abstract}

\section{Introduction}
\label{sec:introduction}
Question Answering (QA) is a central task in Natural Language Understanding (NLU), with numerous industrial applications such as searching information in large corpus, extracting information from conversations or form filling.
Amongst this domain, Reading Comprehension has gained a lot of traction in the past years thanks to two main factors. First, the release of numerous datasets such as SQuAD1.1 \citep{rajpurkar-etal-2016-squad}, SQuAD2.0 \citep{rajpurkar-squad-v2}, BoolQ \citep{boolq}, CoQA \citep{CoQA}, Natural Questions \citep{kwiatkowski2019natural}, only to cite a few. Second, the progress in Language Modeling with the introduction of transformers model \citep{AttentionIsAllYouNeed}, leveraging self-supervised training on very large text corpus, followed by fine-tuning on a downstream task \citep{bert}. This process has now become a de-facto standard for most of Natural Language Processing (NLP) tasks and also contributed to the progress of state-of-the-art for most of these tasks, including Question Answering.

Whilst most of these recent transformers models are English models, French language models have also been released, in particular CamemBERT \citep{camembert} and FlauBERT \citep{flaubert}, as well as multilingual models such as mBERT \citep{multilingual-bert} or XLM-RoBERTa \citep{xlmr}. These models fostered the state-of-the-art in NLP for French language, allowing to benefit from this large-scale transfer learning mechanism.
However, native French resources for Question Answering remain scarcer than English resources. Nonetheless, in 2020, FQuAD1.1 was introduced by Illuin Technology \citep{fquad}. With 60,000+ question-answer pairs, it enabled the development of QA models surpassing the human performance in the Reading Comprehension task.

Although, the introduction of this resource was a major leap forward for French QA, the obtained models suffered from an important weakness, as they were trained to consistently find an answer to the specified question by reading the associated context. In real-life applications however, it is often the case that asked questions do not have an answer in the associated context. For example, let us imagine that we are building a system designed to automatically fill a form with relevant information from property advertisements. We could be interested in the type of property, its surface area and its number of rooms. By asking questions such as \textit{"How many rooms does the property have?"} to a QA model, we would be able to extract this information. But it would also be important that our model is able to predict when such a question does not find an answer in the provided text advertisement, as this situation often arises.

As FQuAD1.1 contains solely answerable questions, the models trained on this dataset did not learn to determine when a question is unanswerable with the associated context. To overcome this difficulty, we extended FQuAD with unanswerable questions, annotated adversarially, in order to be close to answerable ones.

Our contribution sums as follows:
\begin{itemize}[noitemsep,nolistsep]
    \item We introduce the FQuAD2.0 dataset, which extends FQuAD1.1 with 17,000+ unanswerable questions, hand-crafted to be difficult to distinguish from answerable questions, making it the first French adversarial Question Answering dataset with a grand total of almost 80,000 questions.
    \item We evaluate how models benefit from being trained on adversarial questions to learn when questions are unanswerable. To do so, we fine-tune CamemBERT models of varying sizes (large, base) on the training set of FQuAD2.0 and evaluate it on the development set of FQuAD2.0. We take interest in both the ability of a model to distinguish unanswerable questions from answerable ones, as well as an eventual performance drop in the precision of answers provided for answerable questions, due to the addition of unanswerable questions during training. We also study the impact of the number of adversarial questions used and obtain learning curves for each model.
    \item By using both FQuAD2.0 and SQuAD2.0 datasets, we study how multilingual models fine-tuned solely on question-answer pairs of a single language (English), performs in another language (French). We also take interest in performances of such models trained on both French and English datasets.
\end{itemize}

\section{Related work}
\label{sec:related_work}
In the past few years, several initiatives emerged to promote Reading Comprehension in French. With both the release of large scale Reading Comprehension datasets in English such as SQuAD \citep{rajpurkar-etal-2016-squad, rajpurkar-squad-v2}, and the drastic improvement of Neural Machine Translation with the emergence of the attention mechanism within models architectures \citep{bahdanau2016neural, AttentionIsAllYouNeed}, it was at first the most natural path to try to machine translate such English datasets to French. This is what works such as \citet{french-translated-squad} or \citet{kabbadj2018} experimented. However, translating a Reading Comprehension dataset presents inherent difficulties. Indeed, in some cases, the context translation can reformulate the answer such that it is not possible to match it to the answer translation, making the sample unusable. Specific translation methods to mitigate this difficulty were for example proposed in \citet{french-translated-squad} or \citet{spanishsquad}.
Such translated datasets then enable the training of Question Answering models with French data. However, \citet{fquad} demonstrates that the use of French native datasets such as FQuAD1.1 brings far better models. In addition to FQuAD, another French native Question Answering dataset has been released: PiAF \citep{piaf}. This smaller dataset, complementary to FQuAD1.1, contains 3835 question-answer pairs in its 1.0 version and up to 9225 question-answer pairs in its 1.2 version.

We described in Section \ref{sec:introduction} the interest of adversarial Question Answering datasets for some use cases. This concept was first introduced in \citet{rajpurkar-squad-v2} with the presentation of SQuAD2.0, an English dataset extending SQuAD1.1 \citep{rajpurkar-etal-2016-squad} with over 50,000 unanswerable questions. The purpose of SQuAD2.0 is similar to FQuAD2.0's and the hereby presented work takes some of its roots in its English counterpart.

To the best of our knowledge, there is no other dataset for adversarial Question Answering other than SQuAD2.0 and FQuAD2.0, even though there exist several translated versions of SQuAD2.0 in
Spanish \cite{carrino2019automatic},
Swedish \cite{squad_v2_sv},
Polish\footnote{https://bit.ly/2ZqLLgb},
Dutch\footnote{https://bit.ly/3CHFyee},
among others.
This makes FQuAD2.0 the first non-English native dataset for adversarial QA.

Nonetheless, numerous large-scale Question Answering datasets exist apart from FQuAD2.0 and SQuAD2.0, focusing on different Question Answering paradigms.
Some take interest in QA within conversations: CoQA \citep{CoQA} highlights the difficulties of answering interconnected questions that appear in a conversation, QuAC \citep{quac} provides questions asked during an Information Seeking Dialog, ShARC \citep{saeidi2018interpretation} studies real-world scenarios where follow-up questions are asked to obtain further informations before answering the initial question.
Others take interest in more complex forms of reasoning: HotpotQA \citep{yang2018hotpotqa} contains questions that require reasoning among multiple documents, DROP \citep{dua2019drop} introduces questions requiring discrete reasoning types such as addition, comparison or counting.
MASH-QA \citep{zhu-etal-2020-question} takes interest in questions that have multiple and non-consecutive answers within a document. BoolQ \citep{boolq} highlights the surprising difficulty of answering yes/no questions, while RACE \citep{Race} studies multiple choice Question Answering.
The last paradigm we would like to mention is Open Domain Question Answering where a system is asked to find the answer of a given question among a large corpus. This task is tackled with datasets such as Natural Questions \citep{kwiatkowski2019natural} or TriviaQA \citep{joshi-etal-2017-triviaqa}.

Another line of research we find very interesting is a model ability to learn to jointly tackle several Question Answering tasks. \citet{micheli2021structural} trained RoBERTa models \citep{roberta} on both BoolQ \citep{boolq}, a boolean Question Answering dataset, and SQuAD2.0. To enable the model to be multi-task, a slightly modified RoBERTa architecture is presented.
Another way to obtain multi-task Question Answering models is to use an autoregressive model such as GPT-3 \citep{brown2020language} or T5 \citep{raffel2020exploring}. When provided with a context and a question, these text-to-text models directly output an answer in the form of a sequence of tokens using their decoder. This naturally enables such approaches to generalize beyond Reading Comprehension. For example a boolean question can simply be treated by a text-to-text model by outputing "yes" or "no". \citet{khashabi-etal-2020-unifiedqa} leveraged this concept even further by training a T5 model to tackle four different Question Answering formats: Extractive, Abstractive, Multiple Choice, and Boolean. Their model performs then well accross 20 different QA datasets.
We believe that such models for the French language will soon appear, building on French datasets such as FQuAD2.0.

\section{Dataset collection \& analysis}
\label{sec:dataset_collection}

\begin{figure*}[t!]
    \centering
    \includegraphics[width=\textwidth]{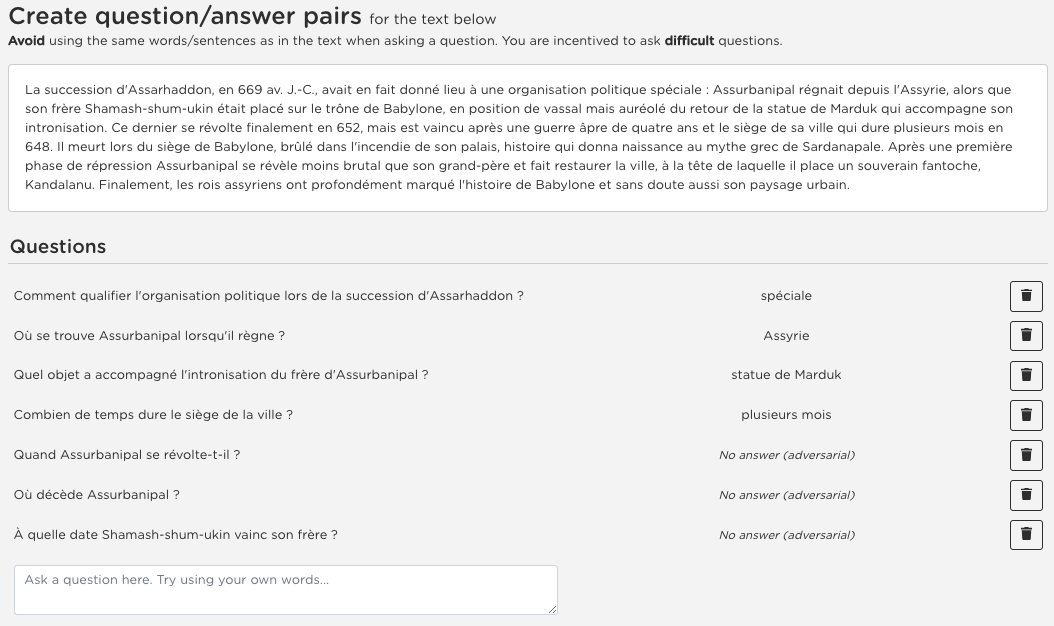}
    \caption{The interface used to collect the question-answer pairs for FQuAD. During the annotation process for FQuAD2.0, an annotator can see a paragraph and the associated answerable questions that were already collected for FQuAD1.1.}
    \label{fig:platform}
\end{figure*}

\subsection{Annotation process}

FQuAD2.0 is an extension of FQuAD1.1 \citep{fquad}.
This extension consists in the addition of unanswerable questions. These questions are hand-crafted in an adversarial manner in order to be difficult to distinguish from answerable ones. To achieve this goal we gave precise guidelines to the annotators:
\begin{itemize}
    \item An adversarial question must be relevant to the context paragraph by addressing a topic also addressed in the context paragraph.
    \item An adversarial question should be designed in the following way: ask an answerable question on the paragraph, and apply to it a transformation such as an entity swap, a negation or something else that renders the question unanswerable.
\end{itemize}

The articles and paragraphs used in the train, development and test sets of FQuAD1.1 and FQuAD2.0 are exactly the same.
An annotator is presented with a paragraph and the already existing answerable questions collected for this paragraph for FQuAD1.1. He is then asked to forge at least 4 adversarial questions, while spending up to 7 minutes by paragraph. A total of 17,765 adversarial questions were collected in 3,100 paragraphs. As FQuAD contains in total 14,908 paragraphs, unanswerable questions were not annotated for every paragraph, nor every article. In order to have reliable evaluations on the development and test sets for this new task, we chose to annotate, in proportion, an important amount of adversarial questions in these two sets. They contain in total around 42\% adversarial questions, while the train set contains 16\% adversarial questions. More statistics can be found in table \ref{tab:fquad2.0}.

We used the Étiquette annotation platform\footnote{\url{https://etiquette.illuin.tech/}} developed by Illuin Technology. It has a dedicated interface to annotate the Question Answering task. Unanswerable questions can be annotated by indicating that the answer is an empty string. A screenshot of the platform is displayed in Figure \ref{fig:platform}.

A total of 18 French students contributed to the annotation of the dataset. They were hired in collaboration with the Junior Enterprise of CentraleSupélec\footnote{\url{https://juniorcs.fr/en/}}.
To limit the bias introduced by an annotator's own style of forging adversarial questions, each annotator only contributed to a given subset: train, development or test.

\begin{table*}[t!]
    \centering
    \resizebox{\textwidth}{!}{
    \begin{tabular}{l c c c c c c c c}
    & & \multicolumn{3}{c}{FQuAD1.1} & \hspace{0.3cm} & \multicolumn{3}{c}{FQuAD2.0} \\
    \hline
    & & Train & Development & Test & & Train & Development & Test \\
    \hline
    Articles & & 271 & 30 & 25 & & 271 & 30 & 25 \\
    Paragraphs & & 12,123 & 1,387 & 1,398 & & 12,123 & 1,387 & 1,398 \\
    Answerable questions & & 50,741 & 5,668 & 5,594 & & 50,741 & 5,668 & 5,594 \\
    Unanswerable questions & & 0 & 0 & 0 & & 9,481 & 4,174 & 4,110 \\
    Total questions & & 50,741 & 5,668 & 5,594 & & 60,222 & 9,842 & 9,704 \\
        \hline
    \end{tabular}}
    \caption{Dataset statistics for FQuAD1.1 and FQuAD2.0}
    \label{tab:fquad2.0}
\end{table*}

\begin{table*}[t!]
    \small
    \centering
   \resizebox{\textwidth}{!}{%
    \begin{tabular}{l l l c}
        Reasoning & Description & Example & Frequency \\
        \hline
        \rule{0pt}{0.35in} Antonym & \specialcell{Use of negation or antonym \\ to make the question adversarial.} & \specialcell{Question: Quels \textcolor{blue}{mamifères} \textbf{ne} \textcolor{red}{sont} \textbf{pas} \textcolor{red}{présents} ? \vspace{0.1in}\\ Context: [...] Le parc \textcolor{red}{abrite} aussi de nombreux grands \textcolor{blue}{mammifères} comme \\ \textbf{des ours noirs, des grizzlys}, [...]} \vspace{0.05in}& 21.6 \% \\
        \hline
        \rule{0pt}{0.35 in} Entity Swap & \specialcell{A name, a number, a date has \\ been modified so that the \\ question becomes adversarial.} & \specialcell{Question: Quelle est la couleur \textcolor{red}{traditionnelle} de \textcolor{blue}{la ville de Paris} ? \vspace{0.1in}\\ Context: [...] La livrée des rames est personnalisée, associant \textbf{le vert jade} \\ \textcolor{red}{traditionnel} de \textbf{la RATP} à divers visuels symboliques de \textcolor{blue}{la ville de Paris}.} & 24.5 \vspace{0.05in} \% \\
        \hline
        \rule{0pt}{0.35 in} Ambiguity & \specialcell{A tiny precision or imprecision \\ in the question makes the \\ plausible answer in the context \\ incorrect.} & \specialcell{Question: Quelle est la \textbf{dernière} station de \textcolor{blue}{la ligne} ? \vspace{0.1in}\\ Context: [...] \textcolor{blue}{La ligne} se dirige vers l'est en position axiale jusqu'à \\ \textbf{la station Balard} [...]} & 17.6 \vspace{0.05in} \% \\
        \hline
        \rule{0pt}{0.5 in} Out-of-context & \specialcell{While some concepts of the \\ question are discussed in the \\ context, at least one key \\ concept of the question is not \\ mentioned in the context.} & \specialcell{Question: Quelle était \textbf{la profession} de \textcolor{blue}{Nicolas Bachelier} ? \vspace{0.1in}\\ Context: Les projets les plus réalistes sont présentés au roi au XVIe siècle. \\ Un premier projet est présenté par \textcolor{blue}{Nicolas Bachelier} en 1539 aux États de \\ Languedoc, puis un second en 1598 par Pierre Reneau, et enfin un \\ troisième projet proposé par Bernard Arribat de Béziers en 1617 [...] } \vspace{0.05in}& 6.9 \% \\
        \hline
         \rule{0pt}{0.55 in} Semantical Similarity & \specialcell{All concepts of the question are \\ mentioned in the context, while \\ the question remains unanswered \\ in the context} & \specialcell{Question: Quel est le nom du \textcolor{blue}{troisième volet de la saga} ? \vspace{0.1in}\\  Context: [...] Le fait que Solo soit plongé dans la carbonite constitue en \\ outre une alternative pour les scénaristes si Harrison Ford refuse de jouer \\ dans \textcolor{blue}{le troisième volet de la saga}. En effet, George Lucas n'est pas assuré \\ que sa vedette accepte de reprendre à nouveau le rôle après son succès \\ dans \textbf{Les Aventuriers de l'arche perdue}.} \vspace{0.05in}& 29.4 \% \\
        \hline
    \end{tabular}
    }
    \caption{Categories of adversarial questions and their respective proportion in a FQUAD2.0 sample of 102 questions. Bold words are the plausible answers or discriminative terms within the question. Colored terms are co-references between question and context.}
    \label{tab:dataset-analysis}
\end{table*}

\subsection{Statistics}

To maximize the number of available questions for fine-tuning experiments, while keeping a sufficiently important set for evaluation, we decide to merge the train and test sets of FQuAD2.0 into a bigger training set, and keep the development set intact for evaluating the obtained models. The new training set contains a total of 13,591 unanswerable questions.
Main statistics for FQuAD1.1 and FQuAD2.0 are presented in Table \ref{tab:fquad2.0}.

\subsection{Challenges raised by adversarial questions}
\label{sec:questions-challenges}

To understand what the different types of adversarial questions collected are, we propose a segmentation of the challenges raised by adversarial questions in FQuAD2.0.
To do so, we randomly sampled 102 questions from the new annotated questions in FQuAD2.0 development set and manually inspected them to identify the challenges they proposed. Then, we sorted these questions following the different identified categories, in order to estimate the proportion of each category within the total dataset. Table \ref{tab:dataset-analysis} presents this analysis where 5 main categories have been identified.

\begin{table*}[t!]
    \fontsize{6}{8}\selectfont
    \centering
    \renewcommand{\arraystretch}{1.1}
    \resizebox{\textwidth}{!}{%
    \begin{tabular}{l c c c c c c c}
        \hline
        Model & Dataset & EM & F1 & F1\textsubscript{has ans} & NoAns\textsubscript{F1} & NoAns\textsubscript{P} & NoAns\textsubscript{R} \\
        \hline
        CamemBERT\textsubscript{BASE} & FQuAD2.0 & 63.3 & 68.7 & 82.5 & 62.1 & 82 & 49.9 \\
        CamemBERT\textsubscript{LARGE} & FQuAD2.0 & \textbf{78} & \textbf{83} & \textbf{90.1} & \textbf{82.3} & \textbf{93.6} & \textbf{73.4} \\
        \hline
    \end{tabular}}
    \caption{Baseline results on the FQuAD2.0 validation set while training is made on the expanded training set containing 13,591 unanswerable questions.}
    \label{tab:baselines_scores}
\end{table*}

\section{Evaluation metrics}
\label{sec:eval_metrics}
To evaluate the predictions of a model on the FQuAD2.0 dataset, we use mainly the Exact Match (EM) and F1 score metrics, which are defined exactly as in \citet{rajpurkar-squad-v2} with the required adaptations regarding stop words for the French language as explained in \citet{fquad}.
In a nutshell, EM measures the percentage of predictions matching exactly one of the ground truth answers, while F1 computes the average overlap between the predicted tokens and the ground truth answer.

To extend these metrics to unanswerable metrics, unanswerable questions are simply considered as answerable questions with a ground truth answer being an empty string.

One may be interested in evaluating on the one hand the ability of a model to extract the correct answers of answerable questions, and on the other hand its ability to determine if a question is unanswerable given the context. To do so, we introduce two other metrics:
\begin{itemize}
    \item F1\textsubscript{has ans}: the average F1 score, question-wise, as defined above, but limited to answerable questions,
    \item NoAns\textsubscript{F1}: the F1 score of the classification problem consisting in determining if a question is unanswerable. It is then the harmonic mean of the precision (NoAns\textsubscript{P}) and recall (NoAns\textsubscript{R}) for this classification problem, the no-answer class being considered as the positive class.
\end{itemize}
We must emphasize that NoAns\textsubscript{F1} is a metric computed as a whole on the entirety of the FQuAD2.0 development set as a classification problem, while F1\textsubscript{has ans} is computed question-wise and is an average of the individual scores for each question. We also want to point out that the global F1 score is in no way the weighted average of F1\textsubscript{has ans} and NoAns\textsubscript{F1}.

In order to evaluate the ability of a Question Answering model to learn when a question is unanswerable, we carried out various fine-tuning experiments using the FQuAD2.0 dataset. These experiments are split into the following sections: French Monolingual Experiments and Multilingual Experiments.

\section{French monolingual experiments}
\label{sec:monolingual-experiments}

The goal of these experiments is two-fold. First, we want to obtain strong baselines on the FQuAD2.0 dataset. Second, we want to analyze how the performances of the fine-tuned models evolve with respect to the quantity of unanswerable questions available at training time.

\subsection{Baselines}
\label{sec:baselines}

To fulfill our first goal, we choose to fine-tune CamemBERT models, because they are the best performing models on several NER, NLI and Question Answering French benchmarks \citep{camembert, fquad}. We could also have chosen FlauBERT models \citep{flaubert}, but \cite{fquad} tends to show that for the same size, CamemBERT models outperform FlauBERT models on the Question Answering task, hence our choice of CamemBERT models. We benchmark two different model sizes: CamemBERT\textsubscript{LARGE} (24 layers, 1024 hidden dimensions, 12 attention heads, 340M parameters) and CamemBERT\textsubscript{BASE}  (12 layers, 768 hidden dimensions, 12 attention heads, 110M parameters).

The fine-tuning procedure used is identical to the one described in \citet{bert}, and an implementation can be found in HuggingFace’s Transformers library \citep{Wolf2019HuggingFacesTS}. All models were fine-tuned on 3 epochs, with a warmup ratio of 6\%, a batch size of 16 and a learning rate of $1.5\cdot 10^{-5}$. The optimizer used is AdamW with its default parameters. All experiments were carried out on a single Nvidia V100 16 GB GPU. Whenever necessary, gradient accumulation was used to train with batch size not fitting within the GPU memory. The results obtained on the FQuAD2.0 development set for the different metrics are presented in Table \ref{tab:baselines_scores}.

These first results allow us to draw the following conclusions:
\begin{itemize}
    \item One can see that the best trained model, CamemBERT\textsubscript{LARGE}, obtains a rather high score of 82.3 \% for the NoAns\textsubscript{F1} metric, while keeping a high score of 90.1 \% for the F1\textsubscript{has ans} metric. It confirms that it is possible for a pre-trained French Language Model to learn to determine with high precision when a French question is unanswerable, while extracting the correct answer in most cases when a question is answerable.
    \item As observed in \citet{fquad} or \citet{lepetit}, Question Answering seems to be a complex task for a small size (base, small) fine-tuned Language Model to solve, and hence the obtained performances are highly dependent to model size, bigger models performing much better than smaller ones. It appears that for Adversarial Question Answering this observation is even more important, with CamemBERT\textsubscript{LARGE} scoring a 20.2 \% absolute improvement in NoAns\textsubscript{F1} metric compared to CamemBERT\textsubscript{BASE}.
\end{itemize}

\begin{table}[t!]
    \small
    \centering
    \renewcommand{\arraystretch}{1.3}
    \resizebox{0.47\textwidth}{!}{%
    \begin{tabular}{l c c c c }
        & \multicolumn{2}{c}{FQuAD1.1} & \multicolumn{2}{c}{FQuAD2.0} \\
        \hline
        Model & EM & F1 & EM & F1 \\
        \hline
        CamemBERT\textsubscript{BASE} & 78.1 & 88.1 & 73.1 & 82.5\\
        CamemBERT\textsubscript{LARGE} & 82.4 & 91.8 & 81.3 & 90.1 \\
        \hline
    \end{tabular}}
    \caption{Comparison of scores obtained on the FQuAD1.1 dev set for models trained on FQuAD1.1 or FQuAD2.0.}
    \label{tab:comparison-with-fquad11}
\end{table}

\pgfplotsset{compat = newest, legend style={at={(0.7,0.2)},anchor=west}}

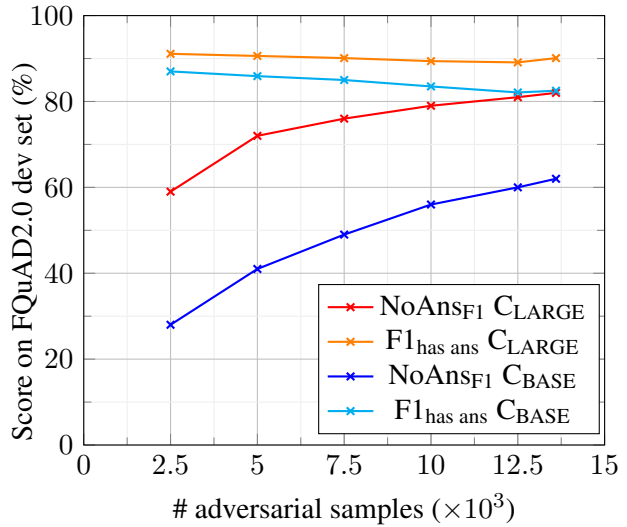
\begin{figure}[t!]
    \centering
    \begin{tikzpicture}[
        trim left=-0.3in, 
        trim right=\columnwidth-0.3in,
        scale=1.0
    ]
        \begin{axis}[
            xmin = 0, 
            xmax = 15,
            ymin = 0, 
            ymax = 100, 
            xtick distance = 2.5,
            ytick distance = 20,
            grid = both,
            minor tick num = 1, 
            major grid style = {lightgray},
            minor grid style = {lightgray!25}, 
            point meta=y,
            xlabel = \# adversarial samples ($\times 10^3$),
            ylabel = Score on FQuAD2.0 dev set (\%),
            ylabel style={at={(-0.07, 0.45)}},
            legend style={at={(0.45, 0.2)}},
        ]
            \addplot[color=red, mark=x, thick] coordinates {
              (2.5, 59)
              (5.0, 72)
              (7.5, 76)
              (10.0, 79)
              (12.5, 81)
              (13.6, 82)
            };
            \addplot[color=orange, mark=x, thick] coordinates {
              (2.5, 91.1)
              (5.0, 90.6)
              (7.5, 90.1)
              (10.0, 89.4)
              (12.5, 89.1)
              (13.6, 90.1)
            };
            \addplot[color=blue, mark=x, thick] coordinates {
              (2.5, 28)
              (5.0, 41)
              (7.5, 49)
              (10.0, 56)
              (12.5, 60)
              (13.6, 62)
            };
            \addplot[color=cyan, mark=x, thick] coordinates {
              (2.5, 87)
              (5.0, 85.9)
              (7.5, 85)
              (10.0, 83.5)
              (12.5, 82.1)
              (13.6, 82.5)
            };
            \legend{NoAns\textsubscript{F1} C\textsubscript{LARGE}, F1\textsubscript{has ans} C\textsubscript{LARGE},
            NoAns\textsubscript{F1} C\textsubscript{BASE}, F1\textsubscript{has ans} C\textsubscript{BASE}}
      \end{axis}
    \end{tikzpicture}
    \caption{Evolution of NoAns\textsubscript{F1} and F1\textsubscript{has ans} for CamemBERT models depending on the number of unanswerable questions in the training dataset}
    \label{fig:learningcurve}
\end{figure}

\subsection{Comparison with FQuAD1.1 scores}

Whilst the models presented in the previous sub-section clearly learned to both extract accurate answers from answerable questions and determine when a question is unanswerable, one may also be interested in whether these models extract as accurate answers as similar models solely fine-tuned on FQuAD1.1, ie. only on answerable questions.

To answer this problematic, we present in Table \ref{tab:comparison-with-fquad11} a comparison of our models of interest in two different set-ups: when fine-tuned solely on FQuAD1.1 and when fine-tuned on the entirety of FQuAD2.0. All evaluations are on FQuAD1.1 dev set. By dataset construction, the F1 score on the FQuAD1.1 dev set is strictly equivalent to the F1\textsubscript{has ans} on the FQuAD2.0 dev set. Results for fine-tuning on FQuAD1.1 are extracted from \citet{fquad}.

With the addition of unanswerable questions during fine-tuning, the model is encouraged to predict that some questions are unanswerable. And as for every model, the NoAns\textsubscript{P} is strictly lower than 100\%, there are answerable questions in the dev set, for which models tend to wrongly predict that they are unanswerable. Then for these questions, the predicted answer is the empty string instead of the expected answer. Hence, we can expect a decrease of the F1\textsubscript{has ans} metric in comparison to the set-up where a model is fine-tuned solely on FQuAD1.1.

This assumption is confirmed in Table \ref{tab:comparison-with-fquad11} with a shrinking gap as model size grows. Indeed, F1\textsubscript{has ans} is only 1.7 absolute points lower for CamemBERT\textsubscript{LARGE} trained on FQuAD2.0 compaired to the same model fine-tuned solely on FQuAD1.1. For CamemBERT\textsubscript{BASE}, the gap grows to 5.6 points.
This gap evolution also follows the evolution of the NoAns\textsubscript{P} metric which is equal to 82\% for CamemBERT\textsubscript{BASE} and 93.5\% for CamemBERT\textsubscript{LARGE}.

\subsection{Learning curves}
\label{sec:learning-curves}

To get a better grasp of how many adversarial questions are needed for a model to learn to determine when a question is unanswerable, we conduct several fine-tuning experiments with an increasing number of adversarial questions used for training.
For every training, all answerable questions of the training set of FQuAD2.0 (i.e. the training set of FQuAD1.1), and unanswerable questions are progressively added to the training set with increments of 2500 questions. We conduct such experiments for the two model architectures of CamemBERT\textsubscript{BASE} and CamemBERT\textsubscript{LARGE}. The results are displayed in Figure \ref{fig:learningcurve}.

\begin{table*}[t!]
    \small
    \centering
    \renewcommand{\arraystretch}{1.3}
    \resizebox{\textwidth}{!}{%
    \begin{tabular}{l c c c c c c}
        \hline
        Model & Training Dataset & Test Dataset & EM & F1 & F1\textsubscript{has ans} & NoAns\textsubscript{F1} \\
        \hline
        \hline
        \multirow{2}{*}{XLM-R\textsubscript{BASE}} & SQuAD2.0 & FQuAD2.0 & 56 & 62.4 & 75.9 & 56.2 \\
        & SQuAD2.0 + FQuAD2.0 & FQuAD2.0 & \textbf{64.4} & \textbf{69.6} & 78.4 & \textbf{66.4} \\
        \hline
        CamemBERT\textsubscript{BASE} & FQuAD2.0 & FQuAD2.0 & 63.3 & 68.7 & \textbf{82.5} & 62.1 \\
        \hline \hline
        CamemBERT\textsubscript{BASE} & FQuAD2.0* & FQuAD2.0* & 60.5 & 66.1 & 83.5 & 56.4 \\    RoBERTa\textsubscript{BASE} & SQuAD2.0* & SQuAD2.0* & 69.7 & 73.3 & 85.3 & 73.4 \\
        \hline
        \hline
        \multirow{2}{*}{XLM-R\textsubscript{LARGE}} & SQuAD2.0 & FQuAD2.0 & 67.3 & 73.4 & 87.8 & 68.1 \\
        & SQuAD2.0 + FQuAD2.0 & FQuAD2.0 & 76.8 & 82.1 & 87.2 & 81.9 \\
        \hline
        CamemBERT\textsubscript{LARGE} & FQuAD2.0 & FQuAD2.0 & \textbf{78} & \textbf{83} & \textbf{90.1} & \textbf{82.3} \\
        \hline
    \end{tabular}}
    \caption{Results for multilingual experiments with FQuAD2.0 and SQuAD2.0. Best score on FQuAD2.0 development set for both model sizes are highlighted in bold. }
    \label{tab:multilingual}
\end{table*}

From these experiments, we observe the following:
\begin{itemize}
    \item The CamemBERT\textsubscript{LARGE} model needs quite few adversarial examples before achieving decent performances. Indeed the model trained with 5k adversarial questions achieves 88\% of the performance of the best model trained with 13.6k adversarial questions, which is 2.7 times more unanswerable questions.
    \item The slope of the CamemBERT\textsubscript{BASE} learning curve is higher than for CamemBERT\textsubscript{LARGE}. For example, the CamemBERT\textsubscript{BASE} model trained with 5k adversarial questions achieves only 66\% of the performance of the best CamemBERT\textsubscript{BASE} model trained with 13.6k adversarial questions. We conclude that the value brought by additional data is more important for smaller models than for bigger ones. However, we also observe that the CamemBERT\textsubscript{LARGE} model trained with 2.5k adversarial questions performs on par with the CamemBERT\textsubscript{BASE} model trained with 12.5k adversarial questions (5 times more data). 
    \item Whatever the model, the learning curve has not flatten yet, which means that both architectures would benefit from more adversarial training samples. In order to do so, one would need to annotate further adversarial questions, which we leave for future work.
\end{itemize}

\subsection{Baseline performances by question category}

We present in section \ref{sec:questions-challenges} a detailed analysis of the different challenges FQuAD2.0 adversarial questions provide. To understand how well the baseline CamemBERT models trained perform on each one of these challenges, we present in table \ref{tab:results-by-category} evaluation results on each of these categories. As the evaluation is solely made on adversarial questions, the chosen metric is the recall of the NoAns task: NoAns\textsubscript{R}.

\begin{table}[t!]
    \small
    \centering
    \renewcommand{\arraystretch}{1.3}
    \resizebox{0.47\textwidth}{!}{%
    \begin{tabular}{l c c}
        & \multicolumn{2}{c}{NoAns\textsubscript{R}} \\
        \hline
        Category & C\textsubscript{BASE} & C\textsubscript{LARGE} \\
        \hline
        Antonym & 36 & 68 \\
        Entity Swap & 36 & 86 \\
        Ambiguity & 50 & 56 \\
        Out-of-context & 43 & 71 \\
        Semantical Similarity & 37 & 70 \\
        \hline
    \end{tabular}}
    \caption{Baseline models recalls for NoAns task on each category of adversarial questions. C\textsubscript{BASE} refers to CamemBERT\textsubscript{BASE}.}
    \label{tab:results-by-category}
\end{table}

\section{Multilingual experiments}
\label{sec:multilingual-experiments}

The previous experiments focus on the study of fine-tuning French Language Models with the FQuAD2.0 dataset. However, one could ask the following questions:
\begin{itemize}
    \item Does a multilingual Language Model fine-tuned solely on English Question Answering datasets could compete against such a model, thanks to the existence of several large-scale English Question Answering datasets?
    \item Does the combination of French and English Question Answering datasets during training makes a multilingual Language Model better than a monolingual Language Model in this Question Answering task?
\end{itemize}

We use FQuAD2.0 and SQuAD2.0, respectively as French and English reference datasets for these experiments. We benchmark two multilingual models: XLM-RoBERTa\textsubscript{BASE} and XLM-RoBERTa\textsubscript{LARGE} \citep{xlmr} which are comparable in size respectively to CamemBERT\textsubscript{BASE} and CamemBERT\textsubscript{LARGE}. Experimental set-up and parameters (relative to model sizes) are identical to the ones described in Section \ref{sec:monolingual-experiments}.
Fine-tuning experiments are summarized in Table \ref{tab:multilingual}.

One can make the following observations from these results:
\begin{itemize}
    \item Results in zero-shot setting are promising. We call zero-shot setting the FQuAD2.0 evaluations of models trained solely on SQuAD2.0 because the models were not trained with any question-answer pair in French. For example, XLM-R\textsubscript{LARGE} reaches in zero-shot setting better performances on the FQuAD2.0 dataset than CamemBERT\textsubscript{BASE} trained on FQuAD2.0. Nevertheless, this observation must be put into perspective by reminding that SQuAD2.0 training set includes 43.5k adversarial questions, hence 3.2 times more than FQuAD2.0. By relying on the learning curves presented in Section \ref{sec:monolingual-experiments}, one can suppose that a CamemBERT\textsubscript{BASE} trained with 43.5k French adversarial questions would have substancially better performances than our actual best CamemBERT\textsubscript{BASE} model.
    \item For both model sizes, the CamemBERT model reaches better performances than the XLM-R model in the zero-shot setting with a substantial margin. One can then conclude than with 13.5k adversarial questions, we are beyond the point where training a French monolingual model on French question-answer pairs brings better results than using a multilingual model trained solely on English question-answer pairs.
    \item By combining FQuAD2.0 and SQuAD2.0 training sets, XLM-R\textsubscript{BASE} performs slightly better than CamemBERT\textsubscript{BASE} trained with FQuAD2.0, while for large models, CamemBERT is slightly better. We believe this is another demonstration of the ability of bigger language models to perform well in small data regimes, while smaller language models need large-scale datasets to reach their potential. Hence, it seems more interesting in a low computing resource setting and low training data availability setting to rely on multilingual models leveraging the more important availability of training data in English.
    \item As a matter of comparison with its English counterpart SQuAD2.0 and in order to assess the relative difficulty of each dataset, we also performed experiments on subsets of SQuAD2.0 and FQuAD2.0, respectively denoted SQuA2.0* and FQuAD2.0*, each with a training set of 50,000 answerable questions and 10,000 unanswerable ones. We trained a RoBERTa\textsubscript{BASE} model and a CamemBERT\textsubscript{BASE} with the same experimental set-ups. The scores obtained for all metrics are significantly better for the English set-up than the French one, notably with increases of 9.2\% for the EM score and 6.2\% for the overall F1 score. Besides, there is a 17\% increase for the NoAns\textsubscript{F1} score for the English set-up. These results clearly indicate that the overall task for FQuAD2.0 is much harder than for SQuAD2.0 in terms of both the difficulty and the ambiguity of questions. In particular, the results for NoAns\textsubscript{F1} indicate that it is much harder for the French model to detect whether a question is answerable.
\end{itemize}


\section{Conclusion \& future work}
\label{sec:conclusion}
In this paper, we introduced FQuAD2.0, a QA dataset with both answerable questions (coming from FQuAD1.1) and 17,000+ newly annotated unanswerable questions, for a total of almost 80,000 questions.
To the best of our knowledge, this is the first French (and, perhaps most importantly, non-English) adversarial Question Answering dataset.

We trained various baseline models using CamemBERT architectures. Our best model, a fine-tuned CamemBERT\textsubscript{LARGE}, reaches 83\% F1 score and 82.3\% F1\textsubscript{no ans}, the latter measuring its ability to distinguish answerable questions from unanswerable ones.
The study of learning curves with respect to the number of samples used for training such models show that our baseline models would benefit from additional unanswerable questions.
In the future, we plan to collect additional samples to expand FQuAD2.0. For comparison, its english cousin SQuAD2.0 \cite{rajpurkar-squad-v2} contains 53,775 unanswerable questions. Such a large-scale dataset would of course enable the acquisition of even better models as the ones presented in Sections \ref{sec:monolingual-experiments} and \ref{sec:multilingual-experiments}.
As far as data collection is concerned, we could also collect additional answers for each unanswerable question. By following the same procedure as in \citet{fquad}, this would allow for the computation of human performance, measuring the inherent difficulty of the challenge provided by FQuAD2.0.

On top of monolingual French experiments, we conducted various multilingual experiments demonstrating the relevance of using multilingual models fine-tuned on English resources to use in other languages when very few or no resources are available in this target language.
Nevertheless, we also showed the superiority of a monolingual approach on the target language using a dataset such as FQuAD2.0 (as this was in our case both economically and practically feasible).
Besides, we also exhibited that FQuAD2.0 is probably a harder dataset of adversarial Question Answering than its English counterpart SQuAD2.0, as similar models perform better in the same conditions in English than in French.

Although the performances that we obtained on the FQuAD2.0 dataset are very good, the evaluation of the resulting models on other datasets is of crucial importance.
In real-life industrial use cases, the contexts and questions asked vary from those present in FQuAD2.0: how can we perform efficient domain transfer on these datasets? This will be further evaluated in future iterations of this work.

Another topic of interest in the context of industrial use cases is the inference time of such large models. The best model we obtained is a CamemBERT\textsubscript{LARGE}, but in some real-life applications where GPUs are unavailable or when we must handle a large number of requests in a short amount of time, we cannot afford the inference times that come with such large models.
The use of smaller models than CamemBERT\textsubscript{LARGE} or CamemBERT\textsubscript{BASE} comes naturally into mind, such as LePetit \cite{lepetit}, also denoted as CamemBERT\textsubscript{SMALL}.
To overcome this limitation, we could use model compression techniques such as pruning \citep{structuredpruning, movementpruning}, distillation \citep{distillation, tinybert, mobilebert} or quantization \citep{kim2021ibert, qbert}.
We performed preliminary tests that seem very promising, they should be investigated further in the future.

\section*{Acknowledgments}
We are very grateful to Robert Vesoul, CEO of Illuin Technology and Co-Director of CentraleSupélec's Digital Innovation Chair, for enabling and funding this project, while leading it through.

We warmly thank Martin D'Hoffschmidt, who launched this initiative within Illuin Technology, and vastly contributed in the first steps of this journey.
Our thanks also go to Inès Multrier for her contribution and insights on multilingual experiments.

We would also like to thank Enguerran Henniart, Lead Product Manager of Étiquette, for his assistance and technical support during the annotation campaign.

Finally we extend our thanks to the whole Illuin Technology team for their innovative mindsets and incentives, as well as their constructive feedbacks.

\bibliographystyle{acl_natbib}
\bibliography{anthology,paper}

\cleardoublepage

\appendix

\end{document}